\documentclass[a4paper]{IEEEtran}
\IEEEoverridecommandlockouts
\usepackage{amsmath,amssymb,amsfonts}
\usepackage{algorithmic}
\usepackage{graphicx}
\usepackage{textcomp}
\usepackage[table]{xcolor}
\usepackage{subcaption}
\usepackage{caption}
\usepackage{mathptmx}
\usepackage{gensymb}
\usepackage{courier}
\usepackage[paper=a4paper,top=104pt, left=55pt, right=34pt, bottom=55pt]{geometry}
\usepackage{hyperref}
\usepackage[capitalise]{cleveref}
\usepackage[scaled=0.9]{helvet}

\usepackage[sorting=none, maxbibnames=1]{biblatex}
\usepackage{enumitem}
\usepackage{multicol}
\usepackage{booktabs}

\begin{document}
\pagenumbering{gobble}
\title{Features characterizing safe aerial-aquatic robots\\
\thanks{
$^{1}$Aerial Robotics Laboratory, Imperial College London.
\par $^{2}$ Laboratory of Sustainability Robotics, Empa - Swiss Federal Laboratories for Materials Science and Technology
\par $^{3}$ eAviation group, Department of Aerospace and Geodesy, TUM 
%\par $^{4}$ EPFL - École Polytechnique Fédérale de Lausanne, Switzerland\\
}
}
\author{Andrea Giordano$^{1,2}$, Luca Romanello$^{2,3}$, Diego Perez Gonzalez$^{3}$, Mirko Kovac$^{1,2}$ and Sophie F. Armanini$^{3}$ }

\maketitle

\begin{abstract}

    This paper underscores the importance of environmental monitoring, and specifically of freshwater ecosystems, which play a critical role in sustaining life and global economy. Despite their importance, insufficient data availability prevents a comprehensive understanding of these ecosystems, thereby impeding informed decision-making concerning their preservation. Aerial-aquatic robots are identified as effective tools for freshwater sensing, offering rapid deployment and avoiding the need of using ships and manned teams. 
    To advance the field of aerial aquatic robots, this paper conducts a comprehensive review of air-water transitions focusing on the water entry strategy of existing prototypes. This analysis also highlights the safety risks associated with each transition and proposes a set of design requirements relating to robots' tasks, mission objectives, and safety measures. To further explore the proposed design requirements, we present a novel robot with VTOL capability, enabling seamless air water transitions.
\end{abstract}

\begin{IEEEkeywords}
    aerial-aquatic robots, robot's design
\end{IEEEkeywords}

\section{Introduction} \label{chap:intro}

%The idea of a flying and submersible robot came to light in the early 1900s and in the following years was discussed in military environments all over the globe. Around 1920, a so-called "flying submarine" was reported by the British journal "Flying" to be the object of conversation of prominent military leaders \cite{flyingsubbritish}. In 1930s, the Russian engineer Boris Ushakov designed the first prototype of flying submarine, which unfortunately came to a halt due to its excessive weight \cite{russian}. However, the fusion between the stealthiness of a submarine and the speed of an aircraft has stimulated the imagination of many, and reappeared years later in technical literature by means of the "Flying submarine" patent filed by Donald Reid in 1958 \cite{patent}. It is only in recent times that the ambition of an aerial aquatic duality gave birth to concrete vehicles. The reason lies in the possibility to create unmanned vehicles, thus much smaller, lighter, and requiring fewer safety precautions. Nowadays, most of the aerial aquatic vehicles found in literature were not built for military applications, but rather for freshwater sensing, search and rescue, and their mission range is of second interest. A total electric propulsion with more modest size of their battery packs, further reducing weight and surface area to ease flight and underwater navigation. 
% Quanto sopra nel paper di giornale

\subsection{Environmental monitoring}

Environmental monitoring is an effective tool in the fight against climate change. In particular, freshwater environments are crucial for life and play a key role in the world economy. The largest freshwater reserves are concentrated in a few spots on the planet's surface, and their monitoring benefits our fight against climate change and helps us prevent potential natural disasters. Freshwater reserves carry substantial economic value, estimated at over $\$4000/ha/year$ ($/ha$ stands for \textit{hectar}) and surpassing that of marine environments by a factor 10, \cite{41e6cea89f2848e5bf561df8fcc0d188}. \newline 
However, data scarcity limits our ability to study and preserve these environments: the scientific community lacks fundamental information about more than 50\% of rivers and lakes, impeding informed decision-making \cite{articleHB}. Moreover, accessing remote or shoreless areas represents an additional challenge in collecting comprehensive data, enhancing the risks for unreachable and sensitive ecosystems. Examples are the coral reef, the polar cap, alpine lakes, etc. 
\newline Difficulties in data collection threaten our ability to meet the UN Sustainable Development Goals (SDGs), especially those related to clean water and sanitation (\# 6), climate action (\# 13), and life below water (\# 14) \cite{COSTANZA2016350}. Addressing these challenges is imperative for precise valuation and the sustainable management of global freshwater environments.
Thanks to their rapid deployment, these aerial aquatic robots emerge as an efficient way of addressing freshwater environmental sensing. They allow access to remote aquatic locations without using ships and thus significantly reducing the costs for data collection and employment of manned teams. Aerial aquatic robots are unmanned, yet not fully automatized: in the future, full automation will augment the benefits they bring to the aquatic environments monitoring.

\subsection{Aerial aquatic applications}

To date, several robots have been designed to implement aerial aquatic applications. In this regard, the focus of developers has been the effectiveness of the aerial aquatic transition and the preservation of the platform itself. On the other hand, the prevention of any ecosystem disturbance resulting from air-water transitions has played a secondary role.
\newline More than a decade ago, when the first prototypes of unmanned aerial-aquatic vehicles came about, the goal was to understand what types of robots could be made aerial-aquatic. This was done through extensive prototyping and on-site testing, which drove the aerial aquatic field through the many possibilities of design, actuation, and control of hybrid robots. After a decade of exploration and despite significant development, platforms are still unable to bridge the gap between research and application reliability. Should this prototyping methodology remain dominant through the next years, aerial aquatic robotics would risk stagnation and would not take root in a broader scientific community.
\newline In this context, we propose a new standard in the design and prototyping of aerial-aquatic robots, specifically including environment-related requirements, safety limitations, and addressing the main challenges. After surveying the state of the art in \cref{chap:relwork}, we propose a new set of standards defining \textit{safe} aerial-aquatic robots in \cref{chap:design} and demonstrate the proposed design approach on a novel test prototype in \cref{chap:robot}.
Conclusively, a prospective analysis of forthcoming research directions is presented in \cref{chap:outlook}.

\section{Related work} \label{chap:relwork}

\begin{figure}[htbp]
    \centering
    \includegraphics[width=1\linewidth]{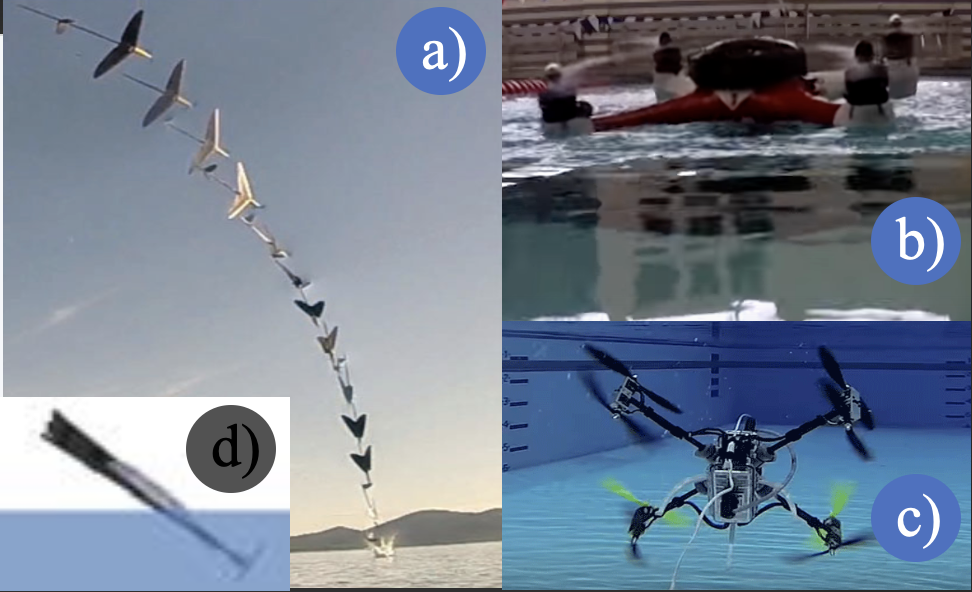}
    \caption{Collage of the drones discussed in \cref{chap:relwork}. a) Plunge diving with AquaMAV: composite image with progressive wing folding. b) Loon Copter during seamless transition. c) Naviator navigating underwater. d) Delta wing by Moore et al. during water entry.}
    \label{fig:drones}
\end{figure}

In the last 15 years, interest in unmanned aerial-aquatic vehicles has risen significantly, giving birth to several new robots. This section discusses the different air-water transition strategies, with a focus on possible damages on the aquatic ecosystems. 
Underwater Aerial Aquatic Vehicles (UAAVs) were developed as a subcategory of Unmanned Aerial Vehicles (UAVs), implementing new features to deal with phase transition. Often, the requirements of aerial flight are in contrast with those of underwater navigation. For instance, the former favours low density, while the latter prefers neutral buoyancy. The challenge in designing effective UAAVs lies in finding suitable compromises between the aerial and the aquatic navigation, to reach satisfying dual locomotion. Additionally, waterproofing, underwater telecommunication and positioning are some of the many other concerns that hybrid robots need to address. 
The aerial locomotion mode influences the type of air-water transition UAAVs undertake. Classified into plunge-diving, seamless and belly landing, transitions are briefly analyzed with emphasis on safety and energy efficiency.

\subsection{Plunge-diving transitions}

The fixed-wing architecture, which implies enhanced speeds and range but reduced stability, does not allow a gradual approach to the water surface, due to stalling risks. Consequently, fixed-wing UAAVs approach the water surface through a nosedive and transition forcefully. To mitigate the collateral effects of such impactful water entries, also addressed as \textit{plunge-diving transitions}, most fixed-wing UAAVs like AquaMAV \cite{siddall2014launching} or Dipper \cite{rockenbauer2021dipper} fold their wings instants before touching the water surface.
Due to the abrupt change in the kinematic viscosity of the medium (at ambient conditions roughly a factor of $15$), robots that plunge dive experience an intense deceleration and severe mechanical stress, leading to potential deformation and critical damage. These collateral effects are caused by the high amount of mechanical energy dissipation over time. Therefore, the key to safe plunge dives is extending the duration of the collisions with water, allowing a gradual rise of the drag after penetration. An effective way to ensure a progressive rise of the slamming force during impacts on water is to have the projectile to resemble the shape of a cone. According to \cite{pandey2022slamming}, in that case the hydrodynamic force $F$ is expressed by the following:

\begin{equation}
    F=\pi \rho V^4 \tan^3{(\frac{\beta}{2} t^2)} 
\end{equation}

where $\rho$ is the water density, $V$ the UAAV's velocity at impact (which for hydrodynamic shapes remains almost unvaried during the impulse), and $\beta$ the cone apex angle. 
The environmental risks deriving from this transition are linked to the possibility of losing the robot or some of its components in the impact. Plausible consequences involve chemical contamination and dispersion of microplastics in water, with damage to the entire ecosystem. Unfavourable weather conditions increase the risk of going off trajectory, resulting in additional slamming forces during plunge diving transitions. This is also the case when the impact on water is not parallel to the drone's velocity vector \cite{pandey2022slamming}. Also, even when running smoothly, plunge dive transitions involve severe decelerations, which can lead to sensor damage and erroneous measurements being carried out. For these reasons, such transitions are not recommended when equipping the robot with delicate payload. 

\subsection{Multicopter UAAVs and seamless transitions}

The multi-copter architecture ensures stability and speed control in vertical motion, allowing the drone to approach the water surface at lower speeds. This architecture favors a smoother type of transition, addressed as seamless, characterized by a longer time scale than plunge diving and a modest amount of energy dissipation. Loon Copter \cite{alzu2018loon} and Naviator \cite{maia2017demonstrating} are two of the multi copter hybrid robots found in literature. The former is a quadcopter featuring a ballast system to ease underwater locomotion. The latter has the shape of a traditional quadcopter drone, but features two coaxial counter-rotating propellers per arm. This octo-quadcopter configuration allows very smooth air-water transitions: when four of the eight propellers approach the air-water interface, their rotation is inhibited and thrust is provided by the remaining blades, independently on which fluid they are immersed in. 
The environmental risks deriving from this transition are milder than those relative to plunge diving, and relate mainly to noise and disturbance caused by the robot's propellers, scaring  off local fauna. Due to noise, this transition affects the data collection process. 

\subsection{Belly landing}
Belly landing is adopted mainly by seaplanes, rather than drones. As the former employ their landing gear to cushion their impact on water, swans and other waterfowls use their feet to damp the contact forces and slow down. As for robots, implementing a landing gear on a small-scale UAAV would substantially increase mass and the control strategy complexity. Despite landing on the belly is an animal inspired strategy to approach the water surface, without diving, it can certainly come in handy for environmental sensing.    
While untested, SUWAVE \cite{7762760} and Moore et al. \cite{8461240} lay the groundwork for executing belly landings. The automation of this method still poses some unsolved challenges concerning attitude control, and the repercussions of water impact on the robot are still uncertain.

\section{Design methodologies} \label{chap:design}

\begin{figure}[htbp]
    \centering
    \includegraphics[width=\linewidth]{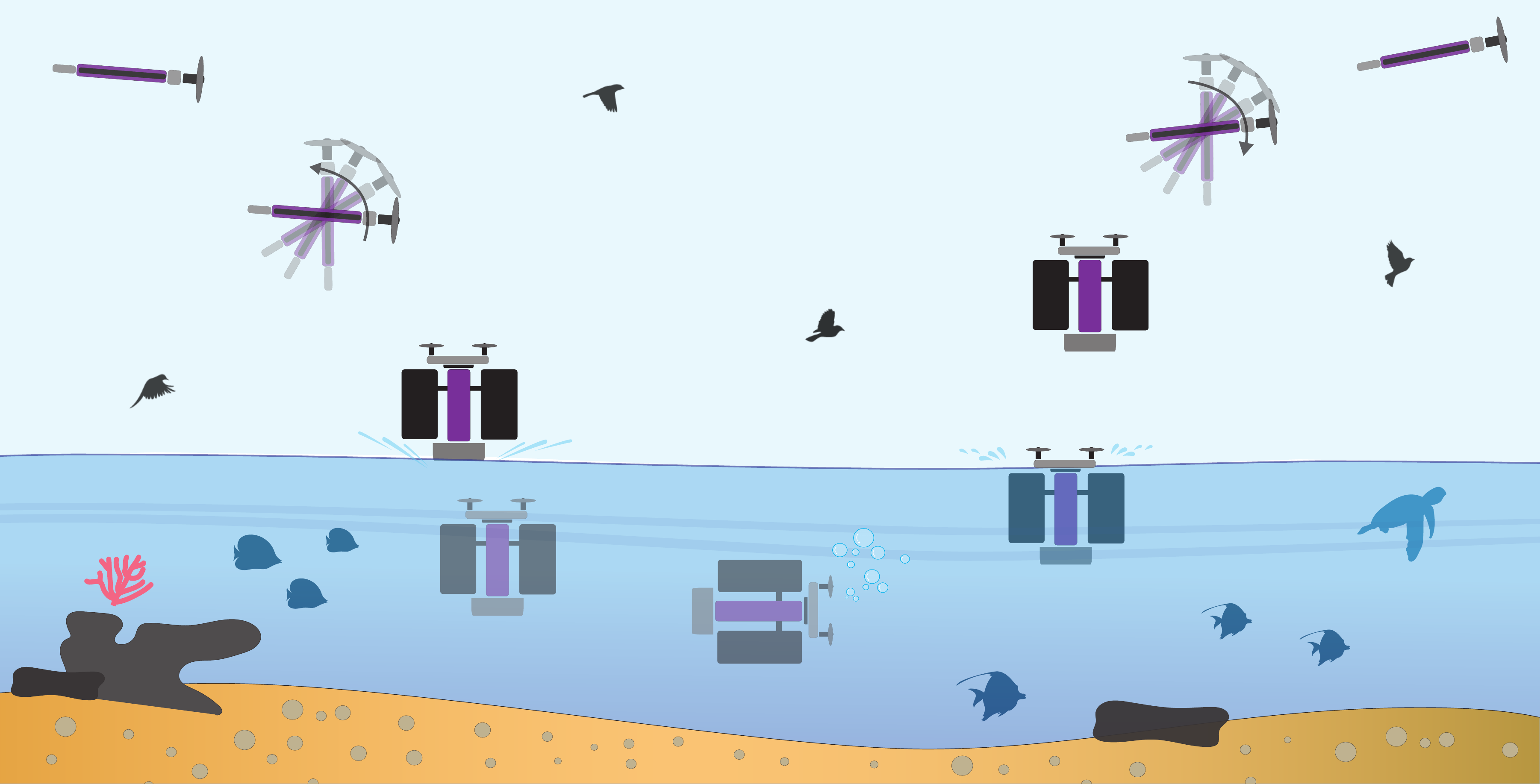}
    \caption{Air-water transitions for a new standard of aerial-aquatic robots}
    \label{fig:mission_complete}
\end{figure}

High-speed diving poses risks to both the platform and the surrounding ecosystem during aerial-aquatic missions. Potential damage upon water impact and the impact on local species need careful consideration. Furthermore, rapid water entry during diving missions may compromise sensing data accuracy and harm water ecosystems. The lack of control in water landing maneuvers prevents full mission autonomy and GPS-based planning. The objective is to establish a design paradigm ensuring environmental and platform safety and efficient controllability for aerial-aquatic robots. These robots are envisioned for deployment in single or multiple lakes, conducting multi-waypoint underwater or water surface missions, incorporating fixed wing flight, hovering, diving, and water takeoff capabilities.
Outlined below are essential design criteria aimed at ensuring the safety and effectiveness of aerial-aquatic robots:

\begin{description}
    \item [Efficiency] to minimize the weight and implement energy-saving aerial and aquatic locomotion modes.
    \item [Safety] to minimize the ecological impact on aquatic ecosystems, with an emphasis on achieving smooth takeoffs and landings on water surfaces. 
    \item [Embodied Automation] to facilitate the automation of a mission through the robot's design. 
    \item [Adaptability] to address control challenges related to the robot's behaviour in diverse locomotion modes. This may involve incorporating adaptable and multi-modal morphologies to accommodate task variations or changes in the medium.
\end{description}

One viable solution that integrates these design criteria is to combine the hovering capabilities of multirotor UAVs, providing high controllability and akin to automation advantages, with fixed-wing flight/ An example is depicted in \cref{fig:mission_complete}. The use of fixed-wing flight proves particularly advantageous for efficiently handling long-range missions with distances ranging from 2 km to 10 km, a characteristic well-suited for the specific demands of freshwater environments \cite{acoustic}. As an example, lake Baikal in Russia is the largest freshwater reserve on the planet, containing alone one fifth of the total freshwater available. Its chemical concentration is constantly monitored and the possibility to conduct long range missions to sample water safely and with viable economic costs would certainly benefit the research in the field \cite{khodzher2016methods}.   
Specifically, we have chosen to adopt the feature of vertical takeoff and landing (VTOL) and to implement it using thrust vectoring. This particular attribute has been extensively explored in the application of UAVs, as discussed in \cite{DUCARD2021107035}. This chosen approach proves beneficial for safety considerations during water entry and exit, ensuring a seamless process. Additionally, it contributes to enhanced automation capabilities. The fixed-wing robot can be initiated without the need for manual hand launch, thereby improving operational efficiency and user comfort. Finally, the robot is adept at smoothly entering and taking off from water, as illustrated in \cref{fig:mission_complete}.

\section{Robot implementation} \label{chap:robot}

\begin{figure}[htbp]
    \centering
    \includegraphics[width=0.9\linewidth]{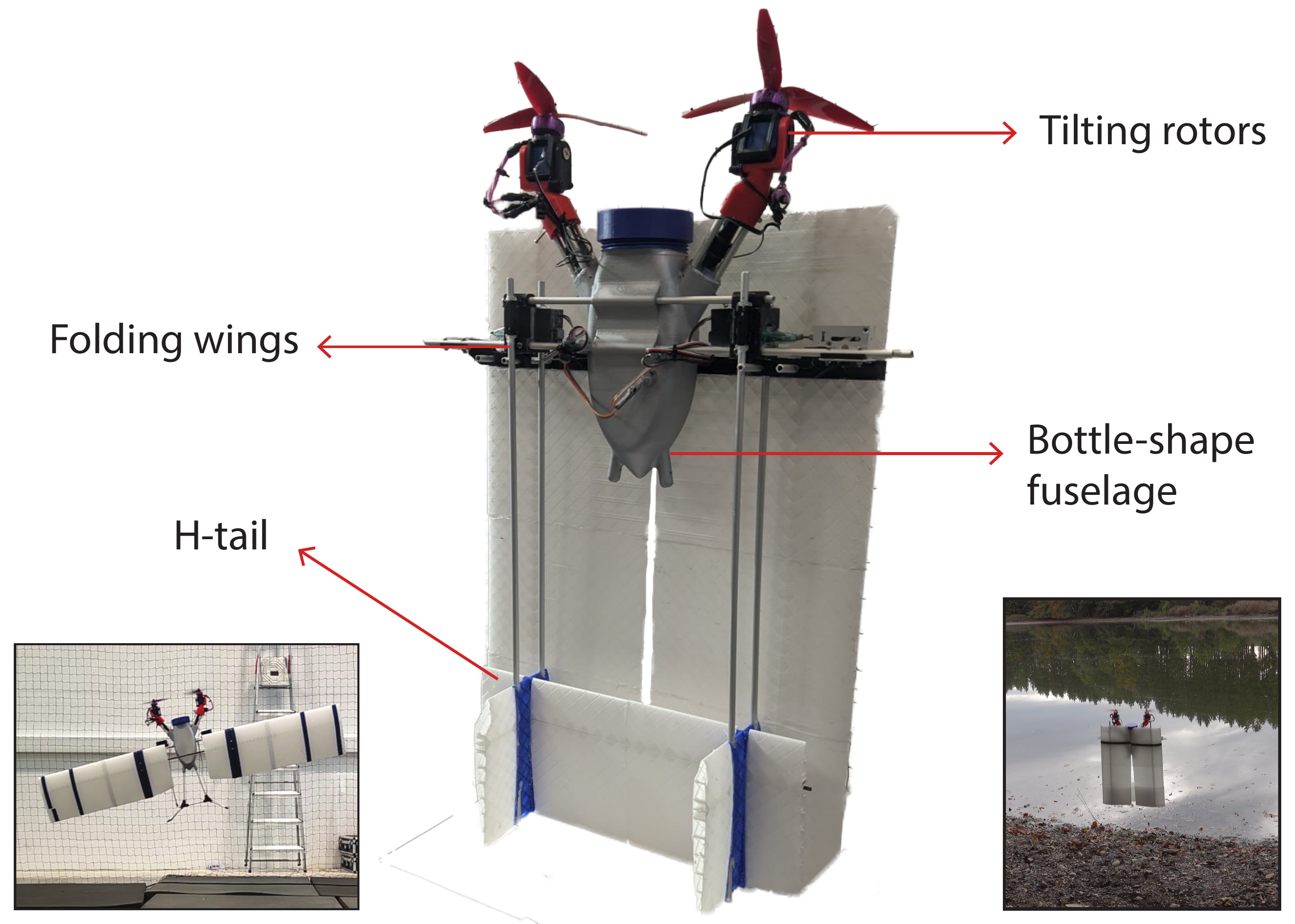}
    \caption{Novel aerial-aquatic prototype with highlighted features and hovering tests}
    \label{fig:aquamav}
\end{figure}
In order to address the design features mentioned above and to achieve VTOL capability, we designed a novel robot prototype (\cref{fig:aquamav}) which incorporates the following design choices:
\begin{description}
    \item[Bi-Copter frame] with two tilting rotors for control in both fixed-wing and hovering modes, minimizing redundancy between water and air propulsion. This design eliminates the need for control surfaces, as demonstrated in Gemini \cite{9001153}. 
    \item[Folding wings] to minimize the unwanted hydrodynamics forces during hovering and underwater operations.
    \item[Tail-sitter] with an H-tail to obtain VTOL capability, enhancing the ground take-off and landing capability 
    \item[Bottle-shaped fuselage] to enhance hydrodynamics characteristic and compactness implemented through thrust vectoring
\end{description}

As depicted in \cref{fig:aquamav}, this platform has undergone the testing phase in hovering mode, with promising performance.

\section{Outlook} \label{chap:outlook}
%The paper's outlook involves the refinement of the control mechanisms for the robot during fixed-wing flight and the seamless transition between hovering and fixed-wing modes, both ways.

%The subsequent phases involve the execution of a complete mission utilizing the presented prototype, with a primary focus on evaluating its capacity to safely and efficiently collect samples from diverse ecosystems. Subsequently, a high-level control system will be designed, leveraging energy consumption data. This system will determine the optimal task to undertake, capitalizing on the inherently embodied automation features of the robot.

An exploration into additional safety considerations will encompass the examination of chemical hazards, necessitating the formulation of design requirements to mitigate potential risks. This investigation will also address issues related to noise and the release of poisonous substances.  Moreover, it will involve assessing the safety scale of utilizing soft robotics, considering their potential to mitigate hazards associated with collision impacts.

To complete our research, the presented prototype will be employed in aerial aquatic missions comprising both water entry and exit, up to reaching satisfying reliability. Primary focus will be on evaluating its ability to collect water samples without endangering the environments, as done in \cref{chap:relwork} for the discussed transitions.
Subsequently, a high-level control system will be designed  to collect energy consumption data of the very platform. This will provide insights on the efficiency of the different locomotions involved, allowing us to safely conduct long range aerial aquatic missions with multiple samplings. 
%As an example, given a complex aerial aquatic scenario where different locations have to be explored, knowing whether it is energetically efficient to take off and fly between every aquatic sampling point or to navigate underwater between some of them, might allow us to complete the mission at all. 
%This system will determine the optimal task to undertake, capitalizing on the inherently embodied automation features of the robot.

\printbibliography[]

\end{document}